# Deep Inferential Spatial-Temporal Network for Forecasting Air Pollution Concentrations

**Hao Wang[1]**
University of Delaware, USA
haowang228@gmail.com

**Bojin Zhuang, Yang Chen**
PingAn Technology, China
{ZHUANGBOJIN232, CHENYANG430}@pingan.com.cn

**Ni Li[1]**
South China University of Technology, China
lini20010@163.com

**Dongxia Wei**
University of Delaware, USA
wdx@udel.edu

[1]The paper was done when these authors were interns in PingAn Technology, Shenzhen, China

## Abstract

Air pollution poses a serious threat to human health as well as economic development around the world. To meet the increasing demand for accurate predictions for air pollutions, we proposed a Deep Inferential Spatial-Temporal Network to deal with the complicated non-linear spatial and temporal correlations. We forecast three air pollutants (i.e., PM2.5, PM10 and $O_3$) of monitoring stations over the next 48 hours, using a hybrid deep learning model consists of inferential predictor (inference for regions without air pollution readings), spatial predictor (capturing spatial correlations using CNN) and temporal predictor (capturing temporal relationship using sequence-to-sequence model with simplified attention mechanism). Our proposed model considers historical air pollution records and historical meteorological data. We evaluate our model on a large-scale dataset containing air pollution records of 35 monitoring stations and grid meteorological data in Beijing, China. Our model outperforms other state-of-art methods in terms of SMAPE and RMSE.

## Introduction

People are paying increasing attention to air pollution over the past decades, since it significantly impacts human health and causes serious harm to other living organisms such as animals and food crops. A lot of air pollution monitoring stations have been built around the world to inform people about air pollution status, such as the concentration of PM2.5, PM10, and $O_3$. Besides monitoring, there is an urgent demand for air pollution prediction, which could benefit the government's policy-making and public outdoor activities(2001).

However, there are huge challenges in making air pollution predictions. First, multiple factors could affect air pollutants and we are not able to capture all of them. For example, pollutants could be generated by various sources such as vehicles emission, factory emissions, and other human activities(Vardoulakis et al., 2003). Unfortunately, it is impossible to gather these data in a real-time manner for a specific location. Although weather plays an important role in the distribution of air pollutants, e.g., strength and direction of wind could affect the diffusion of air pollutants while rain could aid the dissipation of certain pollutants, it is difficult to accurately obtain the future weather information. The failure of taking all the factors into account could cause inaccuracy in the final prediction.

Second, the trends of air pollution concentrations are complicated. The change of different pollutants over time are significantly different. For instance, $O_3$ exhibits quite different behavior from PM2.5 and PM10 on Nov. 14, 2017, at around 2 am as shown in Figure 1 (a). In addition, the same pollutant could behave differently in various regions. In Figure 1 (b), we plot the concentration of PM2.5 as a function of time collected from three air pollution monitoring stations. Wanliu station exhibits a much higher PM2.5 concentration than the other two stations on Jul. 15, 2017 at around 5 am, while slightly lower PM2.5 concentration than Aotizhongxin station on Jul. 17, 2017 at around 1 pm. These complexities make it very difficult for a single model to forecast different air pollutions at variant locations.

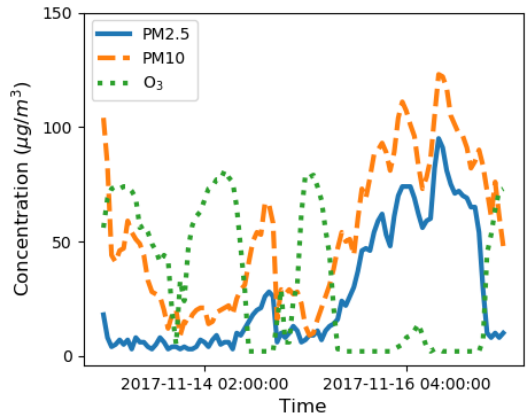

(a) Concentration of different air pollutants

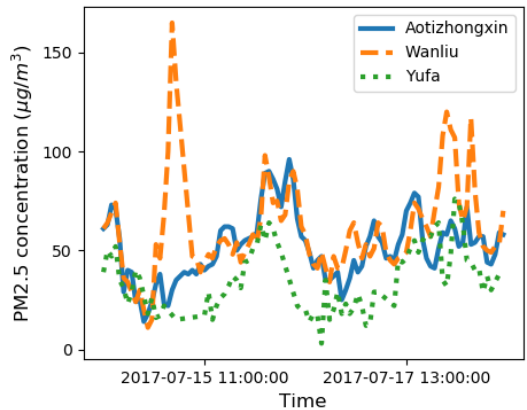

(b) PM2.5 concentration of different stations

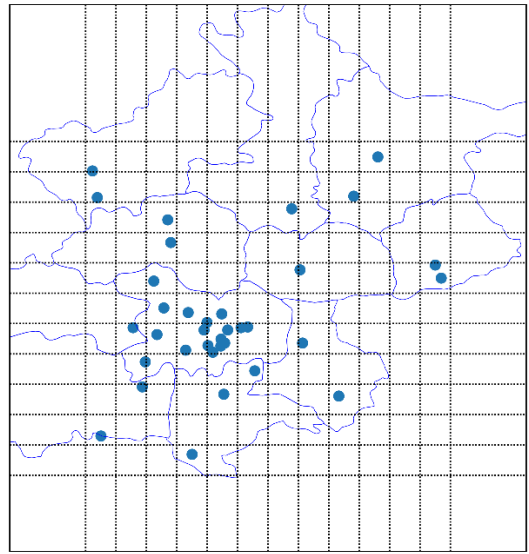

(c) Air pollution monitoring stations in Beijing

Figure 1. (a) Concentrations of different air pollutants changing over time. (b) PM2.5 concentration of three stations changing over time. (c) 35 air pollution monitoring stations in Beijing.

Third, the number of monitoring stations is limited in a city due to huge land usage, expensive maintenance fee and necessary human labor. The data we derived came from only 35 monitoring stations in Beijing. Figure 1 (c) shows the distribution of the air pollution monitoring stations. It is obvious that air pollution monitoring stations present complicated topology and uneven distribution. This makes the prediction even more difficult since the air pollution in a certain region is closely related to that of surrounding areas. The lack of interaction between various regions could lead to inaccurate predictions.

In this paper, we use CNN and sequence-to-sequence jointly to predict the concentration of PM2.5, PM10, and $O_3$ in the future based on the air pollution and meteorology data from the past. As we mentioned above, one challenge of using CNN on spatial is the incomplete grid dataset due to limited number of air pollution monitoring stations. To address this issue, we employ an inferential predictor to fill air pollution time series for regions without a monitoring station. After that, spatial information containing air pollution and meteorological features is treated as an image and handled by CNN while the temporal correlation is captured by sequence-to-sequence architecture with simplified attention mechanism. Additionally, we conduct extensive experiments to compare our proposed deep inferential spatial-temporal network (DIST-Net) with other state-of-art methods and demonstrate superior performance. We can summarize our contributions as follows:

- We proposed a hybrid model that predicts future air pollution with limited data. The model contains inferential predictor, spatial predictor, and temporal predictor to dynamically capture the complicated spatial-temporal relationships.

- To deal with the sparse geographical distribution of air pollution monitoring stations, inferential predictor in our DIST-Net is applied to fill the air pollution of regions with no real data.

- A CNN model over the whole grid is used based on inferential prediction to capture the local characteristics in relation to their neighbors.

- We employed a simplified attention mechanism in temporal predictor. The prediction shows stable performance over the future time steps.

- We trained our model on a large-scale air pollution and meteorological dataset. The result shows that our model outperforms the competing baselines.

## Related Work

Environmental scientists usually predict air pollution based on classical dispersion models, such as Gaussian Plume models, Operational Street Canyon models, and Computational Fluid models(Vardoulakis et al., 2003). These models take the input of meteorology, geometry, emission and traffic factors, and calculate the prediction based on empirical assumption or parameters. However, most inputs mentioned above are difficult to obtain precisely and the parameters in these models may not be applicable to all situations(Vardoulakis et al., 2003).

Some data-driven approaches are also used in the past decades. One class of traditional methods is time series prediction, such as historical average and autoregressive integrated moving average (ARIMA)(Box and Pierce, 1970). But these methods ignore the influence of other features such as meteorology information. Some statistic models, such as linear regression and regression tree(Burrows et al., 1995), have been employed in atmospheric science to do air pollution prediction.

Recently, the success of deep learning in the fields of computer vision and natural language processing motivates the utilization of deep learning methods in air pollution problems(Donnelly et al., 2015, Zhang et al., 2012a, 2012b). For modeling sequential dependency, LSTM network and sequence-to-sequence have been applied for forecasting air pollution. Nonetheless, these models do not consider the spatial correlations. Due to the geographical sparsity of air pollution records, capturing spatial correlation has been a challenge. Zheng et al.(Zheng et al., 2015) divided the surrounding space into 24 regions by three circles and eight angles. Then they fed the aggregated features in each region into the model. Liang et al.(Liang et al., 2018) applied a global spatial attention to capture the readings of other sensors. However, these methods are generally highly engineered compared to CNN.

In contrast to works of literature, our proposed DIST-Net uses inferential predictor to fill the regions without air pollution readings. Therefore, we could harness the power of CNN and GRU for capturing both spatial and temporal relations in a joint deep learning model.

## Preliminaries

We fix some notations and define the air pollution prediction problem in this section.

### Grid

We partitioned a city into $M \times N$ disjointed grids based on the longitude and latitude coordinates, assuming the air pollutions and meteorological features are uniform in a grid. For air pollutions, each grid has a set of air pollutants $P^g = \{p^{g,1}, p^{g,2}, ..., p^{g,m}\}$ to be inferred or already associated if there is an air quality monitor station located. Here, $m$ denotes the type of pollutants. For meteorological features, each grid has a set of weather features $W^g = \{w^{g,1}, w^{g,2}, ..., w^{g,n}\}$ already provided. Here, $n$ denotes the type of weather feature.

Supposing there are $R$ air pollution monitors, each of which generates $m$ kinds of time series corresponding to the concentration of different air pollutants, $P^r = \{p^{r,1}, p^{r,2}, ..., p^{r,m}\}$, for the regions with real pollution monitor, the air pollution features are the time series generated by the air pollution monitor (i.e., $P^g = P^r$). To deal with the region without air pollution monitoring station, we use cubic interpolation to generate a set of time series for different air pollutants, $P^g = \{p^{g,1}, p^{g,2}, ..., p^{g,m}\}$.

### Problem Statement

Given the $\mathcal{T}$-hour time series of target air pollution $j$ and previous time series of all the weather features for the whole grid, denoted as: $X^j = (p^j, w^1, w^2, ..., w^n) \in \mathbb{R}^{T \times M \times N \times (1+n)}$, where $M \times N$ is the size of the grid, the problem is to predict the target air pollution $j$ of gird $i$, over the next $\tau$ hours, denoted as: $Y^{i,j} = (y^{i,j}_{T+1}, y^{i,j}_{T+2}, ..., y^{i,j}_{T+\tau}) \in \mathbb{R}^\tau$.

## DIST-Net Framework

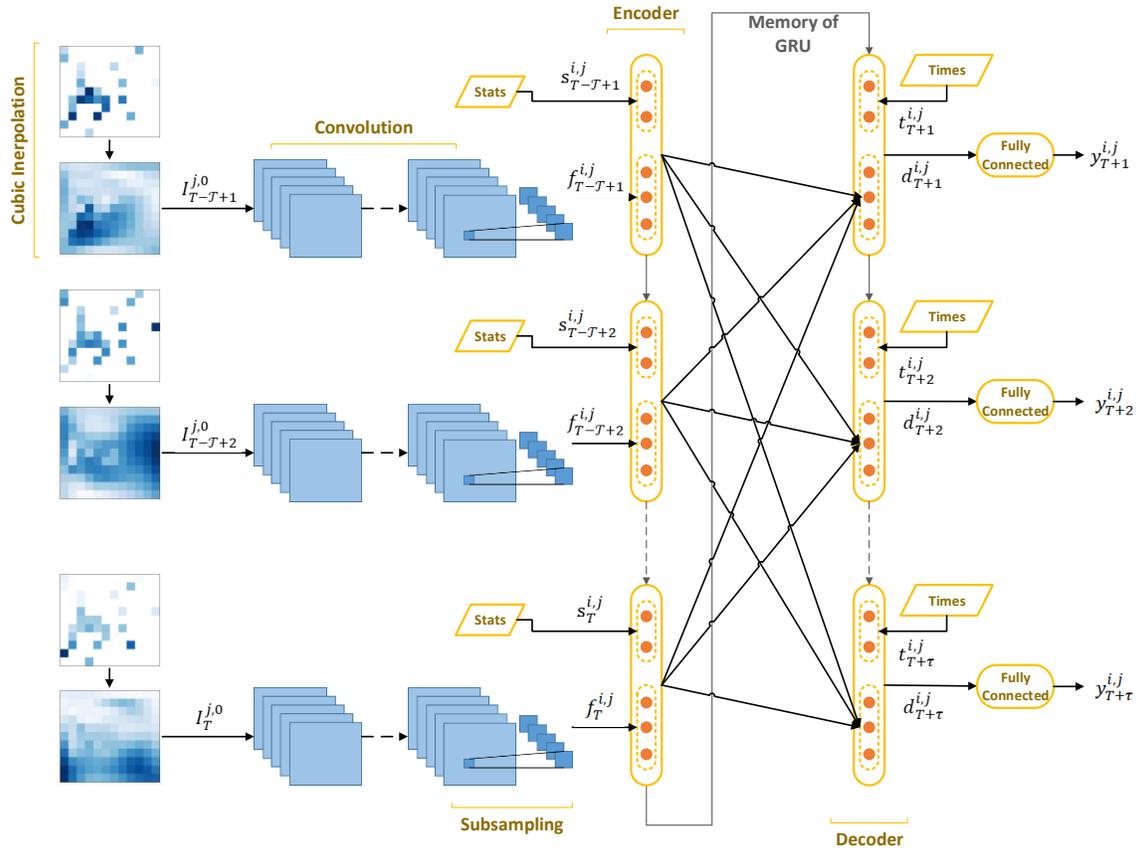

Figure 2. The architecture of DIST-Net. From left to right: inferential predictor fills missing air pollution at regions without air pollution monitoring station by cubic interpolation; spatial predictor utilizes CNN to capture spatial dependency over the whole grid; temporal predictor employs sequence-to-sequence architecture with simplified attention mechanism to generate final predictions.

In this section, the details of our proposed Deep Inferential Spatial-Temporal Network (DIST-Net) is described. The architecture of DIST-Net is depicted in Figure 2. Our model has three components: inferential predictor, spatial predictor, and temporal predictor.

### Inferential Predictor: Cubic Interpolation

Considering the sparse distribution of air pollution monitoring stations, only limited number of regions have real air pollution data while the meteorology data is available on the whole grid. In order to use CNN on spatial sparse data, we use cubic interpolation for the regions with missing data. Here, cubic interpolation only considers neighbor values which is consistent with the First Law of Geography(Tobler, 1970): "near things are more related than distant things".

## Spatial Predictor: CNN

As shown in Figure 2, at each time interval, we feed the whole $M \times N$ grid image into the CNN model. As a result, we have an image as a 3-dimensional tensor $I_t^j \in \mathbb{R}^{M \times N \times (1+n)}$ and $(1+n)$ channels, where 1 is the target air pollution $j$ and $n$ represents the number of weather features. The CNN takes $I_t^j$ at time interval $t$ as input $I_t^{j,0}$ and feeds it into $K$ convolutional layers. The transformation at each layer $k$ is defined as follows:

$$I_t^{j,k} = f(I_t^{j,k-1} * W_t^{i,j,k} + b_t^{i,j,k}) \tag{1}$$

where * denotes the convolutional operation and is an activation function. $W_t^{i,j,k}$ and $b_t^{i,j,k}$ are two sets of learnable parameters in the $k^{th}$ convolution layer. In this paper, we use the scaled exponential linear unit as the activation function. The function is defined as follows:

$$f(x) = \lambda \begin{cases} \alpha(e^x - 1), & x < 0 \\ x, & x \geq 0 \end{cases} \tag{2}$$

where $\alpha = 1.67326$ and $\lambda = 1.0507$.

After $K$ convolution layers, we get a 3-dimensional tensor $I_t^{j,k} \in \mathbb{R}^{M \times N \times \beta}$ at time interval $t$. We pick out one spot, $c_t^{i,j,k} \in \mathbb{R}^{1 \times 1 \times \beta}$, from the tensor where the target station is located, and then use a flatten layer to transform the tensor of picked spot to a feature vector, $f_t^{i,j} \in \mathbb{R}^{\beta}$, as the representation for air pollution $j$ of target station $i$ at time interval $t$.

## Temporal Predictor: Sequence-to-sequence with Simplified Attention

For the temporal view component of the model, an encoder-decoder architecture is proposed to predict the target air pollution of the target grid for the future $\tau$ hours. Gated recurrent units (GRU) has shown similar performance on polyphonic music modeling and speech signal modeling as long short-term memory (LSTM) or even better performance on smaller datasets(Chung et al., 2014). Therefore, we chose GRU as our temporal view component. It was also proposed to address the problem of classic Recurrent Neural Network (RNN) for its exploding or vanishing gradient in the long sequence training(Cho et al., 2014).

In each time interval $t$, the feature vector $f_t^{i,j} \in \mathbb{R}^{\beta}$ are concatenated with statistical analysis $s_t^{i,j} \in \mathbb{R}^{\gamma}$ of air pollution $j$ at station $i$. More specifically, we define:

$$g_t^{i,j} = f_t^{i,j} \oplus s_t^{i,j} \tag{3}$$

where $\oplus$ denotes the concatenation operator. Thus, $g_t^{i,j} \in \mathbb{R}^{\beta+\gamma}$.

Since the performance of sequence-to-sequence architecture will degrade rapidly with increasing encoder length, we propose a simplified attention mechanism to capture all the outputs of the encoder GRU units, $e^{i,j} \in \mathbb{R}^{T \times \delta}$, and then use a fully connected neural network to transform the temporal dimension of the encoder output to the number of decoder units:

$$r^{i,j} = f(W_{att}^{i,j} e^{i,j} + b_{att}^{i,j}) \in \mathbb{R}^{\tau \times \zeta} \tag{4}$$

where $W_{att}^{i,j}$ and $b_{att}^{i,j}$ are both learnable parameters.

Before fed into the decoder GRU units, the transformed encoder output $r^{i,j}$ are then concatenated with embedded time categorical features of the future hours $t^{i,j} \in \mathbb{R}^{\tau \times \eta}$:

$$h^{i,j} = r^{i,j} \oplus t^{i,j} \qquad (5)$$

where $\oplus$ denotes the concatenation operator. Thus, $h^{i,j} \in \mathbb{R}^{\tau \times (\zeta + \eta)}$.

After concatenation, the decoder input $h^{i,j}$ is fed into the decoder GRU with $\tau$ units and the decoder output $d^{i,j}$ is then transformed into the shape of target air pollution concentration series by another fully connected layer:

$$y^{i,j} = f(W_{fc}^{i,j} d^{i,j} + b_{fc}^{i,j}) \in \mathbb{R}^{\tau} \qquad (6)$$

where $W_{fc}^{i,j}$ and $b_{fc}^{i,j}$ are both learnable parameters. $y^{i,j}$ is the final prediction of the proposed DIST-Net model.

**Loss Function**

Our approach is smooth and differentiable. Thus, it can be trained via back-propagation algorithm. Adam optimizer(Kingma and Ba, 2014) is used to train our model by minimize the Symmetric Mean Absolute Percentage Error (SMAPE) between the predicted vector of target air pollution $j$ at station $i$, $y^{i,j}$, and the ground truth vector $Y^{i,j}$:

$$\mathcal{L}(\theta) = \frac{2}{\tau} \sum_{t=1}^{\tau} \frac{|y_t^{i,j} - Y_t^{i,j}|}{|y_t^{i,j} + Y_t^{i,j}|} \qquad (7)$$

where $\theta$ are all learnable parameters in our proposed model.

**Experiment**

**Dataset Description**

This experiment is conducted on a large-scale air pollution monitoring station and mereological station data of Beijing, China between January 1, 2017 and May 31, 2018. We divided Beijing into $11 \times 12$ regions based on 0.1-degree of longitude and latitude coordinates. The size of each region is about $10\ km \times 10\ km$. The air pollution and mereological data are both hourly based. The features include temporal features (e.g., target air pollution and mereological data of the last 72 hours), spatial features (e.g., the coordinates of the air pollution monitoring station within the grid partition), statistical features (e.g., differentials and rolling averages of the previous hours), and time features (e.g., categorical time information of the future 48 hours).

In the experiment, the data from January 1, 2017 to April 30, 2018 is used for training, while the data from May 1, 2018 to May 31, 2018 is used for testing. DIST-Net is designed to predict the future 48 h of target air pollution (e.g., pm2.5, pm10 or $O_3$) at target location based on the air pollution and mereological data of the previous 72 hours.

**Evaluation Metric**

We use SMAPE and Rooted Mean Square Error (RMSE) to evaluate our algorithm. They are defined as follows:

$$SMAPE = \frac{2}{N}\sum_{i=1}^{N}\frac{|y^{i,j}-Y^{i,j}|}{|y^{i,j}+Y^{i,j}|} \qquad (8)$$

$$RMSE = \sqrt{\frac{1}{N}\sum_{i=1}^{N}(y^{i,j}-Y^{i,j})^2} \qquad (9)$$

Where $y^{i,j}$ and $Y^{i,j}$ are prediction value and real value of air pollution $j$ at station $i$. $N$ is the total number of samples.

**Baselines**

We use the following methods to compare with our model. The parameters for all models are tuned for the best performance.

- Autoregressive integrated moving average (ARIMA)(Box and Pierce, 1970): ARIMA is a well-known model for forecasting time series which combines moving average and autoregressive components for modeling time series.

- XGBoost(Chen and Guestrin, 2016): XGBoost is a powerful boosted tree-based method and is widely used for classification or regression purposes.

- Multiple layer perceptron (MLP): This a neural network of two components. The local spatial component is a deep neural network with two fully connect layers of units 32 and 1 respectively on each input time step. The temporal component is another deep neural network with three fully connect layers of units 128, 64 and 48 respectively.

The convolutional neural network is only possible after cubic interpolation on the grid. In order to study the effectiveness of the inferential and spatial predictors, besides the baseline models proposed above, we also devised the following models.

- Local seq2seq: Similar to work of Sutskever et al(Sutskever et al., 2014), this model only contains the temporal part of our DIST-Net model. A seq2seq contains two GRUs: an encoder that processes the input and a decoder that generates the output. For this local seq2seq model, the input only consists of target air pollution, weather, time and statistical features of the target station.

- Neighbor seq2seq: Instead of the inferential and spatial predictor of DIST-Net, we devised spatial partition and spatial aggregation in this neighbor seq2seq model. The spatial partition divides the geographical surrounding space within 10 km radius into eight regions and the spatial aggregation combines the meteorological feature and target air pollution readings located in the same region. This method of capturing spatial correlation is similar to the work of Zheng et al(Zheng et al., 2015). Besides this neighbor spatial predictor, a same seq2seq temporal predictor as in DIST-Net is used to generate predictions.

The same context features and the same loss function (Except for ARIMA) are used for all methods above for fair comparisons.

## Model Comparison

For MLP, local seq2seq, neighbor seq2seq, and DIST-Net models, the input features are normalized to [0, 1] by using Min-Max normalization on the training set. The categorical features (e.g., weather conditions and wind directions) are embedded or one-hot encoded. After evaluation, the predicted values are recovered back to the demand value by applying the inverse of Min-Max transformation.

All these experiments were run on a cluster with two NVIDIA GeForce GTX 1080 Ti GPUs and 64 GB RAM. For MLP, local seq2seq, neighbor seq2seq, and DIST-Net models, the input length is 72 h, while for XGBoost, the input length is 240 h, and for ARIMA, the input sequence is the whole air pollution time series between January 1, 2017 and April 30, 2018.

Table 1: SMAPE and RMSE comparison with different baseline models

| Method | SMAPE | | | RMSE | | |
| --- | --- | --- | --- | --- | --- | --- |
| | PM2.5 | PM10 | $O_3$ | PM2.5 | PM10 | $O_3$ |
| ARIMA | 0.5354 | 0.4862 | 0.8141 | 39.472 | 97.216 | 71.127 |
| XGBoost | 0.5437 | 0.4843 | 0.6790 | 41.132 | 95.668 | 57.299 |
| Multiple layer perception | 0.6348 | 0.7022 | 0.7468 | 206.484 | 606.174 | 109.787 |
| Local seq2seq | 0.5115 | 0.4626 | 0.6638 | 39.544 | 95.777 | 56.164 |
| Neighbor seq2seq | 0.5008 | 0.4691 | 0.6720 | 38.965 | 97.666 | 56.624 |
| DIST-Net | 0.4665 | 0.4189 | 0.6092 | 36.356 | 91.686 | 50.798 |

The above table shows that our DIST-Net model outperforms all other models in both metrics. Comparing to other baselines, for PM2.5, our proposed DIST-Net exhibits at least 6.8% and 6.7% improvements on SMAPE and RMSE respectively, for PM10, at least 9.4% (SMAPE) and 4.3% (RMSE) improvements, and for $O_3$, at least 8.2% (SMAPE) and 9.6% (RMSE) improvements.

## Variant Comparison

## Evaluation on Encoder Length

In Figure 3, we compare the average SMAPE and the average RMSE of different encoder length. The figures show that $O_3$ exhibits relatively constant SMAPE and RMSE values from 24 h to 96 h encoder length. While the average SMAPE of PM2.5 and PM10 from 24 to 72 hours drops 4.3% and 6% respectively. The performance doesn't improve significantly after 72 hours. This could be understood as long historical data has a subtle impact on the future values when compared to the more recent data. Therefore, for our proposed DIST-Net, the past 72 h historical data are chosen to make the predictions.

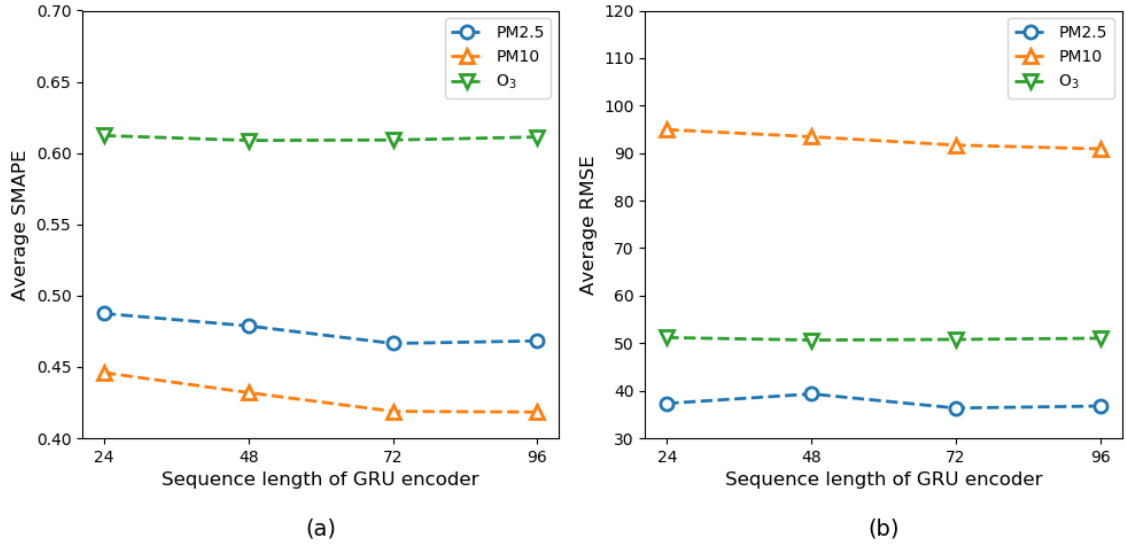

Figure 3. (a) Average SMAPE among all stations with respect to the sequence length of GRU encoder. (b) Average RMSE among all stations with respect to the sequence length of GRU encoder.

**Evaluation on Future Time Step**

We evaluate the performance of DIST-Net on different future time step segments. The length of each segment is set to be 6 h for the purpose of examination. For each segment, we compute the average metric over all air pollution monitoring stations during the corresponding time period. The performances of other commonly used sequence-to-sequence models tend to drop significantly as future time step increases. As illustrated in Figure 4, the performance of our proposed DIST-Net is relatively stable within the 48-hour prediction length. No apparent performance degradation is observed in segmental average SMAPE or RMSE. This demonstrates the stability of our proposed model, especially the success of the sequence-to-sequence architecture with our proposed simplified attention mechanism.

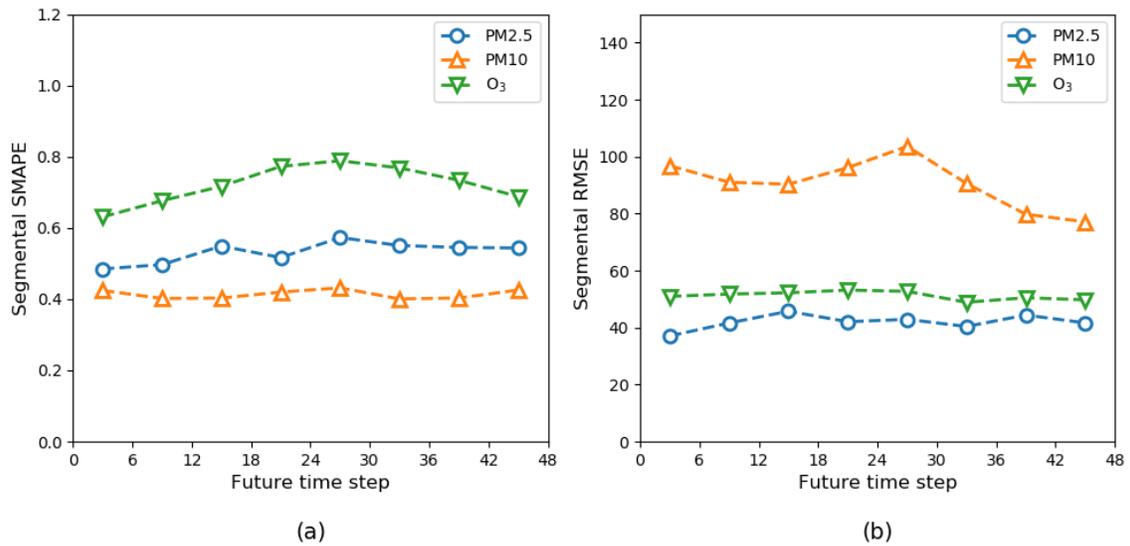

Figure 4. Performance on different segments of future time steps. (a) Segmental average SMAPE among 6 hours vs future time step. (b) Segmental average RMSE among 6 hours vs future time step.

**Predictions against Ground Truth**

The length of prediction is 48 h, as we extract test samples from May 1, 2018 to May 31, 2018 with a slide of 1 h. Thus, each air pollution value of target location at certain time step is predicted for 1 to 48 times. As illustrated in Figure 5, for PM2.5, PM10, and $O_3$, our average predicted values replicate the general trend of the ground truth. Also, most ground truth values lie within the $\mu \pm 2\sigma$ intervals (gray area), where $\mu$ is the average of predictions at each time step and $\sigma$ is the standard deviation.

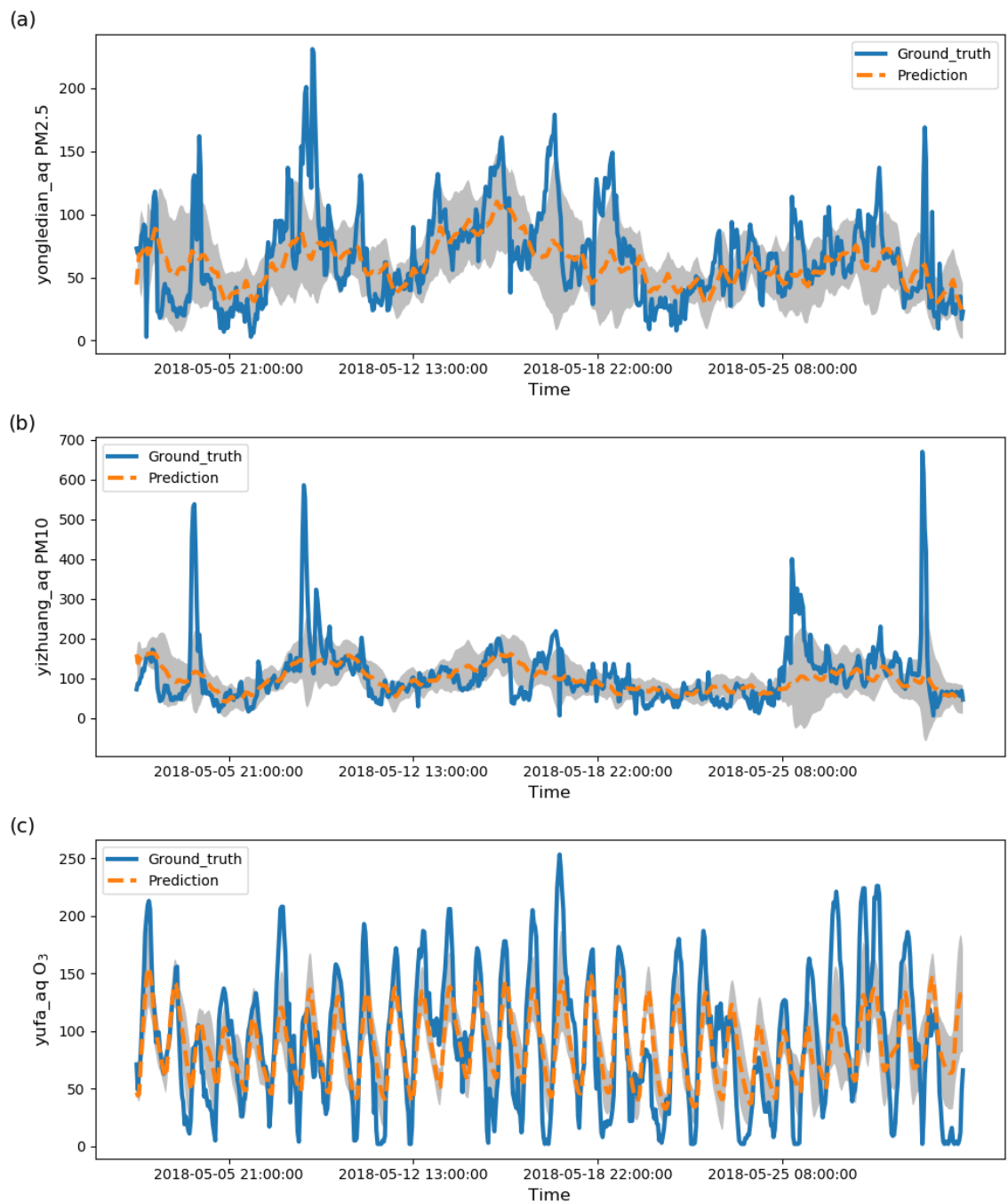

Figure 5. Visualization of air pollution prediction by DIST-Net. The solid blue line is the ground truth; the dashed orange line the average of multiple predictions; the light grey shaded area covers the 2 standard deviations of multiple predictions. (a) PM2.5 of Yongledian air pollution monitoring station. (b) PM10 of Yizhuang air pollution monitoring station. (c) $O_3$ of Yufa air pollution monitoring station.

# Conclusion and Discussion

We proposed a novel DIST-Net model for forecasting the air pollution of a city with limited monitoring stations. Cubic interpolation is employed to address the geographical missing values. Next, CNN is used to capture the spatial dependencies. At last, sequence-to-sequence with simplified attention mechanism is used to model the dynamic temporal correlations. We evaluate our DIST-Net model on a large-scale dataset of air pollution records and meteorological data in Beijing and our DIST-Net outperforms many other baseline models in terms of SMAPE and RMSE.

# Acknowledgments

This work was supported by Smart Environment project of PingAn Technology. The views and conclusions contained in this paper are those of the authors and should not be interpreted as representing any funding agencies.

# References

Air Quality Forecasting: A Review of Federal Programs and Research Needs. Air Quality Research Subcommittee of the Committee on Environment and Natural Resources (CENR); 2001.

Box GEP, Pierce DA. Distribution of Residual Autocorrelations in Autoregressive-Integrated Moving AverageTime Series Models. Journal of the American Statistical Association 1970;65(332):1509-26.

Burrows WR, Benjamin M, Beauchamp S, Lord ER, McCollor D, Thomson B. CART Decision-Tree Statistical Analysis and Prediction of Summer Season Maximum Surface Ozone for the Vancouver, Montreal, and Atlantic Regions of Canada. J Appl Meteor 1995;34:1848-62.

Chen T, Guestrin C. XGBoost: A Scalable Tree Boosting System. KDD 2016. San Francisco, CA, USA2016. p. 785-94.

Cho K, Merrienboer Bv, Gulcehre C, Bahdanau D, Bougares F, Schwenk H, et al. Learning Phrase Representations using RNN Encoder–Decoder for Statistical Machine Translation. EMNLP. Doha, Qatar2014. p. 1724–34.

Chung J, Gülçehre Ç, Cho K, Bengio Y. Empirical Evaluation of Gated Recurrent Neural Networks on Sequence Modeling. NIPS2014.

Donnelly A, Misstear B, Broderick B. Real time air quality forecasting using integrated parametric and non-parametric regression techniques. Atmospheric Environment 2015;103:53-65.

Kingma DP, Ba J. Adam: A Method for Stochastic Optimization. arXiv:14126980 2014.

Liang Y, Ke S, Zhang J, Yi X, Zheng Y. GeoMAN: Multi-level Attention Networks for Geo-sensory Time Series Prediction. IJCAI-182018. p. 3428-34.

Sutskever I, Vinyals O, Le QV. Sequence to Sequence Learning with Neural Networks. arXiv:14093215 2014.

Tobler WR. A Computer Movie Simulating Urban Growth in the Detroit Region. Economic Geography 1970;46:234-40.

Vardoulakis S, Fisher BEA, Pericleous K, Gonzalez-Flesca N. Modelling air qualityin street canyons: a review. Atmospheric Environment 2003;37:155-82.

Zhang Y, Bocquet M, Mallet V, Seigneur C, Baklanov A. Real-time air quality forecasting, part I: History, techniques, and current status. Atmospheric Environment 2012a;60:632-55.

Zhang Y, Bocquet M, Mallet V, Seigneur C, Baklanov A. Real-time air quality forecasting, part II: State of the science, current research needs, and future prospects. Atmospheric Environment 2012b;60:656-76.

Zheng Y, Yi X, Li M, Li R, Shan Z, Chang E, et al. Forecasting Fine-Grained Air Quality Based on Big Data. KDD 2015. Sydney, NSW, Australia2015.